\documentclass{article}

\usepackage[preprint]{draft} %
\usepackage[T1]{fontenc}
\usepackage{amsmath, amssymb}
\usepackage{graphicx}
\usepackage{booktabs}
\usepackage{float}
\usepackage{wrapfig}
\usepackage{placeins}
\usepackage{algorithm}
\usepackage{algpseudocode}
\usepackage{xspace}
\usepackage{algpseudocode}
\usepackage{booktabs}
\usepackage{threeparttable}
\usepackage{graphicx}
\usepackage{makecell}
\usepackage{multirow}
\graphicspath{{figures/}}

\newcommand{\method}{ReCoVLA\xspace}

\title{ReCoVLA: VLM-Guided Reward Compilation for Failure Recovery in Vision-Language-Action Policies}

\title{ReCoVLA: VLM-Guided Reward Compilation for Failure Recovery in Vision-Language-Action Policies}

\author{
  \textbf{Haodi Hu}$^{1,2}$,
  \textbf{Chung-Ta Huang}$^{2,3}$,
  \textbf{Jing Liu}$^2$,
  \textbf{Ye Wang}$^2$ \\
  \textbf{Kei Suzuki}$^2$,
  \textbf{Matthew Brand}$^2$,
  \textbf{Toshiaki Koike-Akino}$^2$ \\
  haodihu@usc.edu \\
  $^1$University of Southern California, USA \\
  $^2$Mitsubishi Electric Research Laboratories (MERL), USA \\
  $^3$Harvard University, USA
}

\begin{document}
\maketitle

\begin{abstract}
    Vision-language-action (VLA) policies provide strong priors for language-conditioned manipulation, but remain brittle in off-nominal states requiring targeted recovery. We propose \method---a failure-conditioned residual recovery framework that keeps a pretrained VLA policy frozen, uses an external vision-language model (VLM) to infer the failure mode and recovery stage, and compiles a structured reward from task-relevant components. Rather than using the VLM to generate actions or rewards directly, \method uses it as a semantic reward selector: it predicts a recovery descriptor and reward mask for in-simulation residual-policy training, followed by zero-shot sim-to-real deployment of the trained recovery policies. This decouples high-level failure understanding from low-level corrective control to support different VLAs. Experiments across short-horizon, long-horizon, and contact-rich manipulation tasks show that \method outperforms the tested baselines on average. In simulation, our reward compiler improves average success from 36.7\% for the fine-tuned $\pi_{0.5}$ baseline to 66.7\%. In physical zero-shot sim-to-real experiments, \method achieves the best average performance, with \(61.7\%\) success.
\end{abstract}


\section{Introduction}
\label{sec:intro}

Large vision-language-action (VLA) policies, especially those with flow matching~\cite{black2024pi_0, intelligence2025pi_05}, have become more capable of mapping language instructions and visual observations directly to robot actions. 
Their scale gives them broad semantic knowledge and strong nominal manipulation skills, but deployment still exposes a practical weakness: imitation-trained policies can enter action-induced states that differ from the training distribution~\cite{ross2011dagger}. 
For instance, when the robot misplaces an object, loses a grasp, or reaches an off-distribution state, one policy that solves the nominal task may lack the failure-specific corrective behavior needed for recovery~\cite{dai2025racer,pan2025selfcorrecting}. 

One direct solution is to fine-tune the VLA policy on additional recovery data. 
However, collecting failure-recovery demonstrations is costly and the fine tuning may forget previously acquired capabilities~\cite{hu2022lora,houlsby2019parameter,kirkpatrick2017overcoming}. 
Reinforcement learning (RL) offers an alternative, but introduces two challenges for VLA recovery. 
First, the reward structure must match the current failure mode: a generic task-level reward is often too sparse~\cite{andrychowicz2017her}, whereas activating every hand-designed reward can introduce conflicting objectives or alter the effective optimization problem~\cite{amodei2016concrete}. 
Second, applying standard policy-gradient RL to flow-matching VLA models is difficult due to lack of likelihood availability.

\begin{figure}[tbp]
    \centering
    \includegraphics[width=\linewidth]{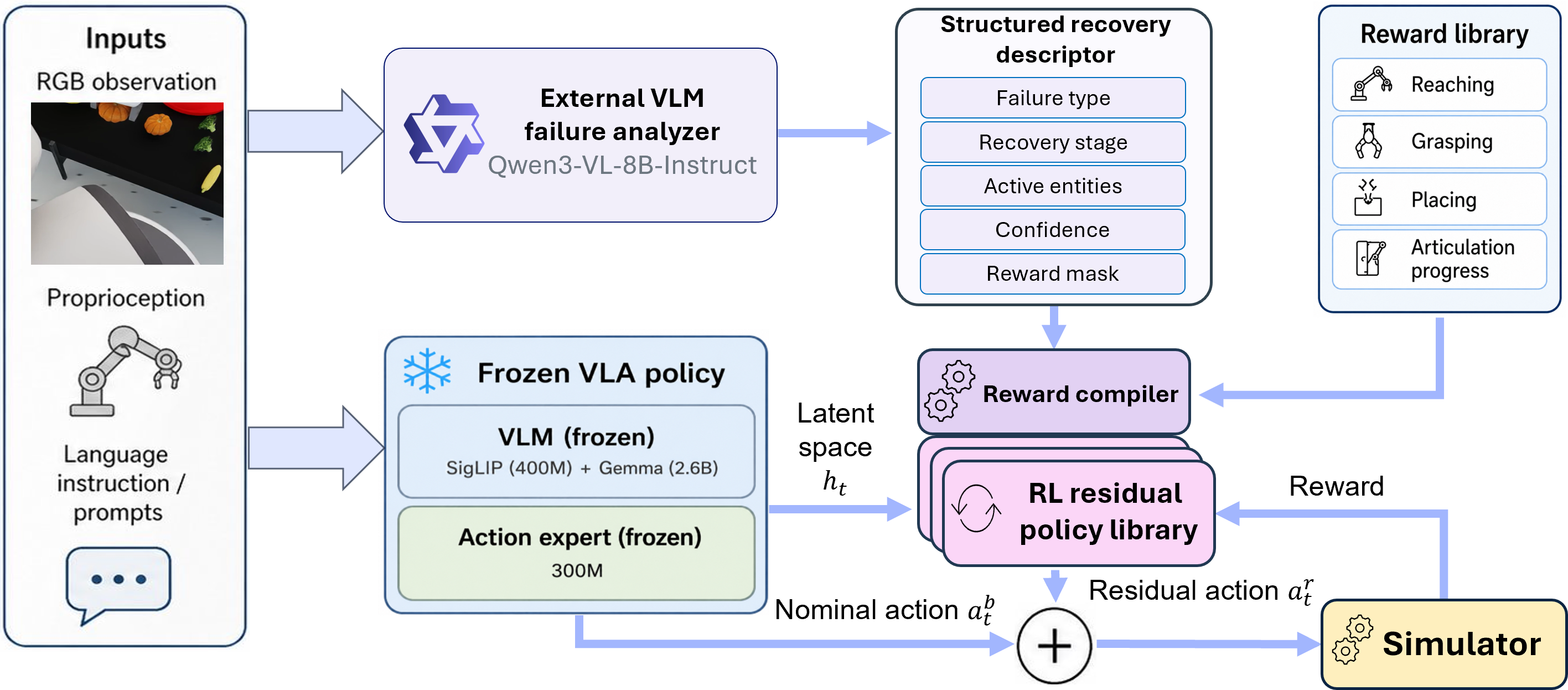}
    \vspace{0 pt}
    \caption{Overview of failure-conditioned residual VLA recovery. The frozen VLA policy maps robot inputs to a nominal action \(a_t^b\) and latent feature \(h_t\). In parallel, an external Qwen3-VL-8B-Instruct VLM analyzes the RGB observation and prompt stream to produce a structured recovery descriptor containing the failure type, recovery stage, active entities, confidence, and reward mask. The reward compiler combines this descriptor with the reward library to generate the reward for residual-RL training. The residual policy observes \(h_t\), receives the compiled reward during training, and outputs a corrective action \(a_t^r\), which is added to the nominal VLA action before execution.}
    \label{fig:method-overview}
    \vspace{-10pt}
\end{figure}

The recent VLA $\pi_{0.6}$~\cite{intelligence2025pi_06} addresses the action-likelihood issue by avoiding direct policy-gradient optimization through offline RL. 
However, it still requires training a separate value function, relies on costly human interventions, and does not directly solve the reward-design problem for failure recovery. 
These challenges motivate a recovery system that keeps the base VLA intact while adapting the reward structure to a diverse set of failure modes. 
As shown in Figure~\ref{fig:method-overview}, ReCoVLA is a
failure-conditioned residual recovery framework that uses an external VLM to analyze off-nominal states and compile a structured reward for residual RL. The VLM does not directly generate robot actions or free-form rewards. Instead, it produces a structured recovery descriptor containing the failure type, recovery stage, active entities, confidence, and reward mask. 
A deterministic reward compiler then grounds the entities, selects reward-library components, and inserts stage gates so that reward terms are active only when their recovery preconditions are satisfied. 
An adaptively selected one of residual policies is then trained over frozen VLA latents, preserving the nominal behavior of the base VLA while learning corrective control for failure states.

We evaluate this design on three Fetch manipulation tasks selected to cover different recovery regimes: long-horizon vegetable sorting, short-horizon soda-can disposal, and contact-rich toolbox organization. 
The experiments show that reward structure is critical for reliable recovery, and
the proposed stage-gated reward compiler achieves the strongest overall performance across baselines. 
In simulation, \method improves average success by 30.0 percentage points over the fine-tuned \(\pi_{0.5}\) baseline. 
In physical Fetch experiments, it also achieves the highest average success, outperforming baselines by 18.3 percentage points. 
These results suggest that semantic failure identification is useful not only for deciding when to recover, but also for determining which reward terms should guide each stage of the recovery behavior.
In summary, our contributions are:
\begin{enumerate}
    \item We introduce a VLM-guided reward compiler that converts structured failure descriptors into executable residual-RL rewards by grounding entities, selecting reward components, and inserting stage gates, instead of using VLM for direct action or reward generation.

    \item We formulate VLA failure recovery as a mixture of residual experts learning over VLA latents, preserving nominal behavior while learning corrective control only for detected off-nominal states.

    \item We evaluate the framework across long-horizon sorting, short-horizon disposal, and contact-rich organization tasks, demonstrating in simulation and on a physical robot that \method outperforms all baselines.
\end{enumerate}

\section{Related Works}

\paragraph{State-of-the-art VLA policies.}
VIMA~\cite{jiang2023vima} and PaLM-E~\cite{driess2023palm} established early VLA frameworks that robot behavior can be conditioned on rich multimodal prompts. 
RT-1/2~\cite{brohan2022rt,brohan2023rt} demonstrated that transformer-based VLA can scale to hundreds of real-world tasks. RoboCat~\cite{bousmalis2023robocat}, RT-X~\cite{openx2023openx}, RoboFlamingo~\cite{li2023roboflamingo}, and GR-1~\cite{wu2023gr1} explored broader pretraining and adaptation. 
More recently, open VLA policies~\cite{ghosh2024octo,black2024pi_0,liu2024robomamba,li2024cogact,qu2025spatialvla,wen2024tinyvla,kim2024openvla,intelligence2025pi_05} highlight several emerging themes like efficient adaptation and improved reasoning. 
Rather than proposing new base VLA, our work focuses on robust recovery when the nominal policy enters failure states.

\paragraph{RL-based refinement.}
Residual reinforcement learning~\cite{johannink2018residualrl}, residual policy learning~\cite{silver2018residualpolicy}, RL token~\cite{xu2026rl}, and residual feedback learning~\cite{ranjbar2021residualfeedback} show that they can repair imperfect controllers more efficiently than pure RL. 
Several works~\cite{gab2024track2act,ankile2024resip,pertsch2020spirl,ball2023rlpd,luo2024serl,luo2024hilserl,zhou2024wsrl,ren2024dppo,fang2025adpo,yuan2024policydecorator,song2025maniplvmr1} study how to adapt pretrained policies with online RL. 
\method is close in spirit to residual refinement, but differs by learning the residual in the latent space of a frozen VLA and by using an external VLM to condition the recovery reward structure on the failure mode.

\paragraph{Failure recovery.}
Prior work has studied damage adaptation and trial-and-error recovery~\cite{cully2015robots}, as well as RL recovery and safety monitors that guide the nominal policy near risky states~\cite{thananjeyan2021recoveryrl}. 
Recent work studies language or VLM-guided recovery through demonstrations, language guidance, or executable actions~\cite{dai2024racer,lin2025failsafe}. 
Instead of relying on manually designed monitors, resets, or task-specific recovery behaviors, \method uses a VLM to produce a structured failure descriptor and compile the residual-RL reward to train mixture of residual policies.

\section{Method}
\label{sec:method}

\paragraph{Overview.}
We propose a failure-conditioned residual recovery framework with three stages. 
First, we execute the frozen base VLA in simulation and use an external VLM to analyze the full failed rollouts, producing a catalog of simulator-observed failure categories and recovery descriptors. 
Second, a deterministic reward compiler constructs a reward from object-state reward components and trains one of  residual policies in simulation for each recoverable failure category. 
Third, during real-world deployment, the VLM monitors a history of physical observations. 
When it detects a known failure category, the system dispatches the corresponding trained residual policy; when the category is unknown, no residual policy is called and the base VLA continues execution. 
The VLM is used for semantic failure identification and recovery-policy selection, not for direct action generation.

\paragraph{Evaluation variants.}
We consider six evaluation variants M1--M6. 
The first group M1--M4 uses the fine-tuned $\pi_{0.5}$ policy as the base model. 
M1 is the frozen $\pi_{0.5}$ base VLA without residual recovery. 
M2 adds a residual policy trained with a task-level reward, in which all task-relevant reward components are active simultaneously. 
M3 is an ablation of \method that uses the VLM-detected failure category to activate only failure-relevant reward components, but applies no stage-aware gates.
M4 is our full proposed method \method, which uses the same failure-relevant reward components as M3, but adds stage-aware gates that activate later-stage terms only after their preconditions are satisfied, reducing premature reward activation and potential reward hacking. 
The second group M5--M6 evaluates whether the same recovery design can be applied to a different VLA backbone. 
M5 is the fine-tuned OpenVLA base policy without residual recovery. 
M6 applies the proposed stage-gated recovery design on top of OpenVLA.

\subsection{Base VLA and residual control}
\label{subsec:base-vla}

We consider a language-conditioned manipulation task with instruction $\ell$. 
At time $t$, the frozen base VLA maps the robot observation, $o_t$, and $\ell$ to a
nominal action and an internal latent feature:
$
(a_t^b,h_t)=F_{\mathrm{VLA}}(o_t,\ell)
$,
where $a_t^b\in\mathcal{A}$ is the base action and $h_t$ is the VLA latent representation used by the residual recovery module. 
The base VLA is kept frozen throughout residual training and deployment. 
A residual policy outputs an additive correction $a_t^r\in\mathcal{A}$ from the VLA latent representation:
$
a_t^r\sim \pi_{\theta}^{r}(\cdot\mid h_t)
$.
When residual recovery is active, the executed action is:
\begin{equation}
\label{eq:residual-action}
a_t
=
\Pi_{\mathcal{A}}
\left(a_t^b+\beta_t a_t^r\right),
\end{equation}
where $\Pi_{\mathcal{A}}$ projects onto the valid action set and $\beta_t\in\{0,1\}$ is a binary deployment-time recovery activation variable. 
This action composition preserves the nominal VLA behavior while allowing the learned residual policy to provide targeted corrections in detected failure states.

\begin{figure}[tbp]
    \centering
    \includegraphics[width=1.0\linewidth]{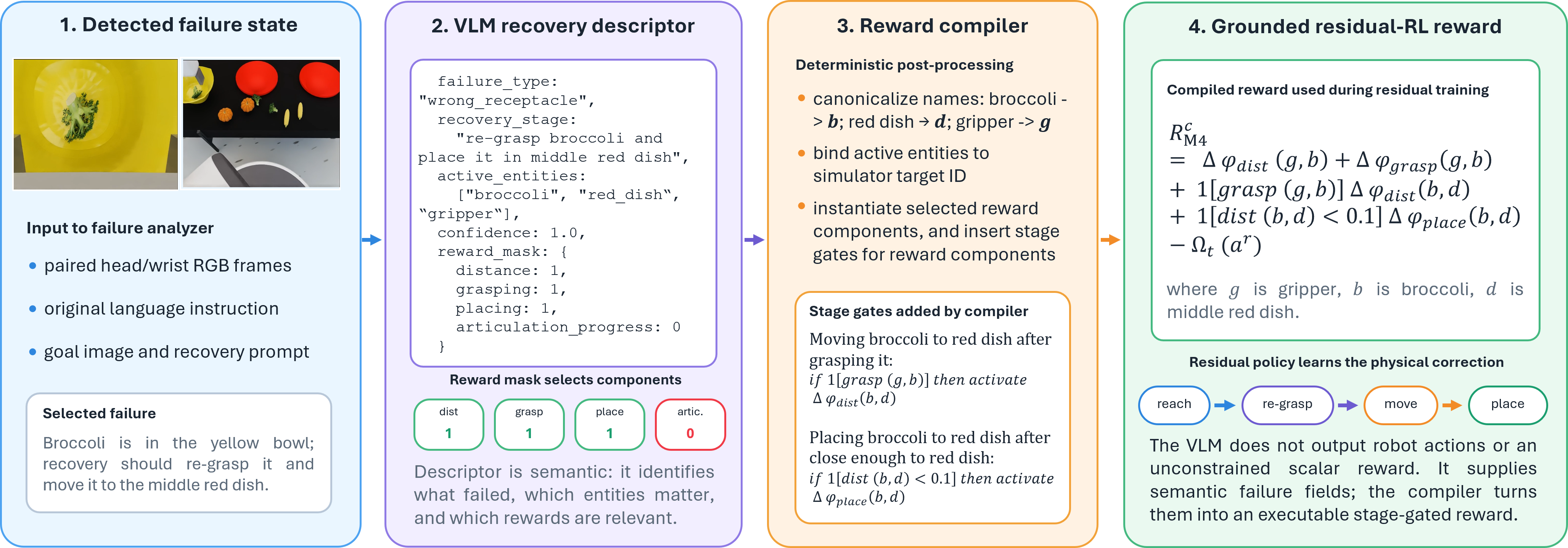}
    \vspace{-10 pt}
    \caption{Example reward-compilation trace. The VLM analyzes the failed rollout and produces a structured recovery descriptor containing the failure type, recovery stage, active entities, confidence, and reward mask. The reward compiler then canonicalizes the entities, binds them to simulator object IDs, selects the corresponding reward-library components, and inserts stage-aware gates.}
    \label{fig:reward-compilation-trace}
    \vspace{-10 pt}
\end{figure}

\subsection{Failure catalog construction}
\label{subsec:failure-catalog}

We construct the set of recoverable failure categories using simulator rollouts from the no-recovery base policy M1. Let $\tau_i^{M1}=\{(o_t, s_t, a_t^b)\}_{t=0}^{T_i}$ denote a rollout generated by the frozen base VLA policy, where $o_t$ is the visual observation, $s_t$ is the robot state, and $a_t^b$ is the base-policy action. For each failed rollout, the external VLM failure analyzer, $D_\phi$, takes as input the full visual rollout, the language instruction $\ell_i$, and a recovery-analysis prompt $q$, and outputs
\begin{equation}
\xi_i = D_\phi(\tau_i^{\mathrm{M1}},\ell_i,q) =
(c_i,z_i,E_i,\rho_i,m_i).
\end{equation}
Here $c_i\in\mathcal{C}$ is the failure category, $z_i\in\mathcal{Z}$ is the recovery stage, $E_i\subseteq\mathcal{E}$ is the set of active entities (e.g., "broccoli", "red\_dish"), $\rho_i\in[0,1]$ is the confidence score for failure states, and $m_i\in\{0,1\}^K$ is a binary mask over the reward-component library. The high-confidence categories form the simulator-trained failure catalog: $\mathcal{C}_{\mathrm{train}} = \{c_i:\rho_i\ge \tau_{\mathrm{cat}}\}$. For each $c\in\mathcal{C}_{\mathrm{train}}$, the descriptor provides the active entities $E_c$ and the failure-relevant reward mask $m_c$ for the reward compiler.

\subsection{Reward variants and stage-gated compilation}
\label{subsec:reward-compiler}

The reward compiler uses a library of $K$ object-state reward potentials $\mathcal{R}=\{\varphi_k\}_{k=1}^{K}$. 
Each potential measures progress for a primitive recovery concept such as reaching, grasping, placing, or articulation. 
For component $k$, we use potential-difference:
\begin{equation}
    \Delta\varphi_k(s_t,s_{t+1};E)
=
\varphi_k(s_{t+1};E)-\varphi_k(s_t;E)
,
\end{equation}

where $E$ denotes the entities used to instantiate the component. 
The residual regularization term is
\begin{equation}
\Omega_t(a^r)=
\lambda_1\|a_t^r\|_2^2+
\lambda_2\|a_t^r-a_{t-1}^r\|_2^2.
\end{equation}
All reward terms are computed from simulator object states. 
The reward variants are defined as follows. 
For M2, the task-level residual baseline activates all task-relevant components and entities for instruction $\ell$. 
For example, in the soda-can disposal task, all reward components related to picking and placing soda cans are enabled. 
Let $m_\ell\in\{0,1\}^K$ denote this task-level mask and $E_\ell$ the set of task-level entities. 
Its reward is
\begin{equation}
\label{eq:m2-reward}
R_{\mathrm{M2}}^{\ell}(s_t,a_t,s_{t+1})
=
\sum_{k=1}^{K}
 m_\ell^{(k)}
 \Delta\varphi_k(s_t,s_{t+1};E_\ell)
-
\Omega_t(a^r).
\end{equation}
For M3, the VLM-detected failure category activates only the failure-relevant components selected by $m_c$, but all selected terms are active simultaneously:
\begin{equation}
\label{eq:m3-reward}
R_{\mathrm{M3}}^{c}(s_t,a_t,s_{t+1})
=
\sum_{k=1}^{K}
 m_c^{(k)}
 \Delta\varphi_k(s_t,s_{t+1};E_c)
-
\Omega_t(a^r).
\end{equation}
M4 uses the same failure mask $m_c$ as M3, but adds a binary stage-aware gate $g_{c,k}(s_t;E_c)\in\{0,1\}$ for each selected component:
\begin{equation}
\label{eq:m4-reward}
R_{\mathrm{M4}}^{c}(s_t,a_t,s_{t+1})
=
\sum_{k=1}^{K}
 m_c^{(k)}
 g_{c,k}(s_t;E_c)
 \Delta\varphi_k(s_t,s_{t+1};E_c)
-
\Omega_t(a^r).
\end{equation}
The gates encode recovery-stage preconditions; M4 differs from M3 only by the stage-aware gates. 
For example, in pick-and-place recovery, the placing component is inactive until the object is grasped, preventing the residual policy from optimizing target motion before it has re-established control of the object. 
M6 uses the same reward as M4, but replaces $\pi_{0.5}$ with OpenVLA.

\paragraph{Reward compiler.}
The reward compiler acts as a deterministic interpreter that maps a schema-valid VLM descriptor into an executable residual-RL reward. 
As illustrated for the sorting task in Figure~\ref{fig:reward-compilation-trace}, the VLM detects a wrong-receptacle placement failure from paired rollout frames and outputs a structured descriptor \(\xi_c=(c,z,E_c,\rho,m_c)\). 
The compiler first verifies that the failure category \(c\) is in the recoverable catalog and that the confidence \(\rho\) and reward mask \(m_c\) satisfy the required thresholds. 
It then canonicalizes the VLM-predicted active entities \(E_c\) using the task specification and the simulator's object map, assigning each entity a semantic role (e.g., \textit{gripper}, \textit{target}, or \textit{articulated part}). Consequently, the active-entity set serves as a candidate pool rather than a rigid argument list. 
Each reward component declares an argument signature to which the compiler binds only the required roles. 
In our sorting example, the descriptor identifies the gripper, the misplaced broccoli, the source yellow bowl, and the target dish; thus, the grasping component is instantiated with \((\text{gripper}, \text{broccoli})\), while the placing component is instantiated with \((\text{broccoli}, \text{target dish})\). 
Next, the compiler attaches fixed stage-gate templates determined by the failure type and recovery stage (\(z\)), such as grasp-before-place or contact-before-articulation. 
This pipeline underscores a critical design choice: the VLM avoids generating unstable, free-form scalar rewards or thresholds. 
Instead, it provides structured semantic fields that the compiler safely grounds into the stage-gated reward used in Eq.~\ref{eq:m4-reward}. 
To validate the stability and reproducibility of these VLM-detected states, we provide a confusion matrix alongside comprehensive compiler implementation details in Appendix~\ref{app:reward-details}.

\begin{figure}[tbp]
    \centering
    \includegraphics[width=\linewidth]{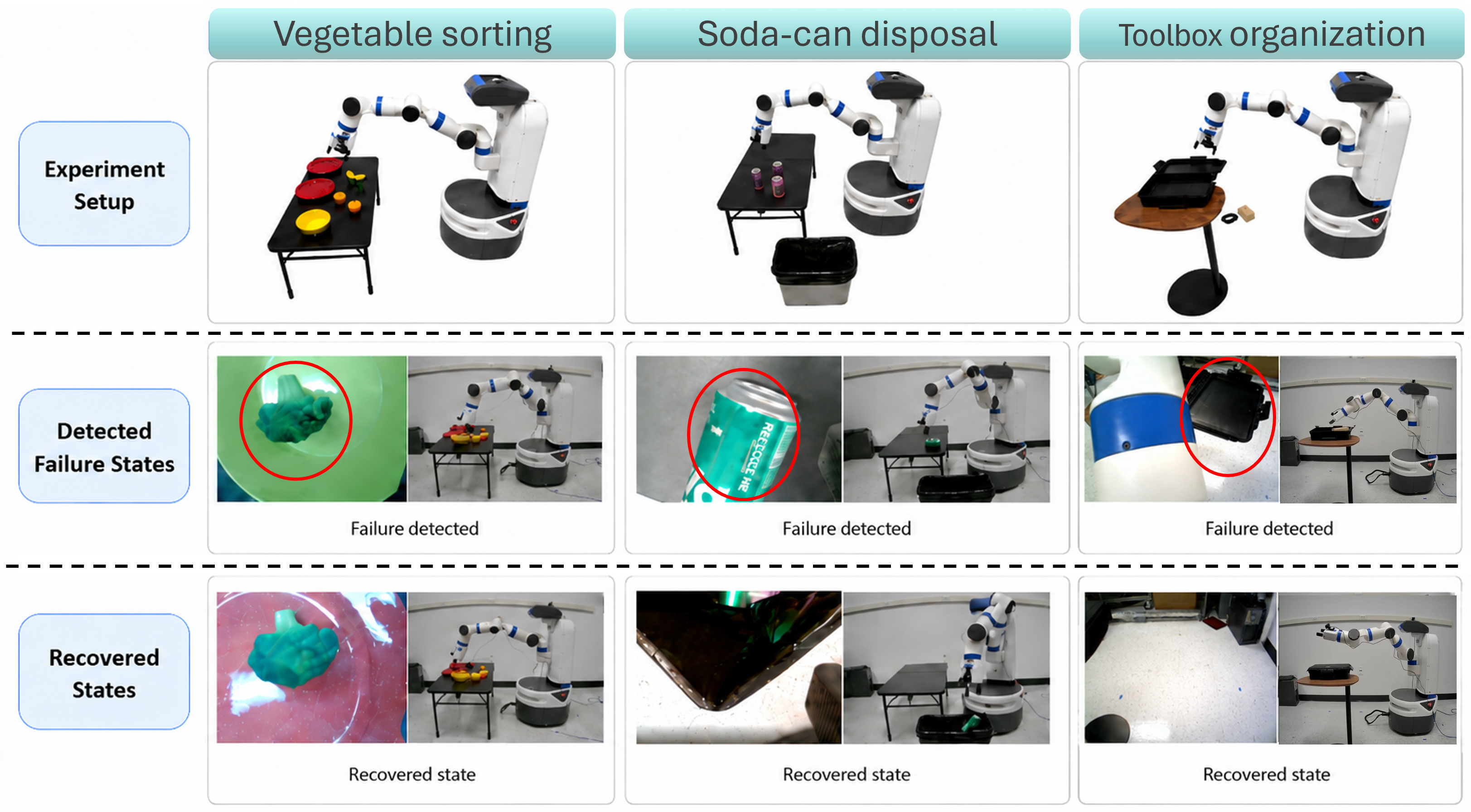}
    \vspace{-10 pt}
    \caption{Physical experiments setup. 
    Columns show the three evaluation tasks: organizing the toolbox; sorting vegetables; and soda-can disposal. 
    Rows show the initial experiment setup, example task progress, a detected failure state from the VLM failure analyzer, and the recovered state after residual correction. 
    The examples illustrate that the recovery descriptor must be task-specific and stage-specific: recovering a dropped cable or box, correcting an object-category placement error, and completing soda-can disposal require different active entities and reward components.}
    \vspace{-6 pt}
    \label{fig:experiment-examples}
\end{figure}

\paragraph{RL residual policy library.} 
Residual policies are trained entirely in simulation. 
For M2, we train a task-level residual policy using $R_{\mathrm{M2}}^{\ell}$. 
For M3, M4, and M6, we train failure-category residual policies using the corresponding rewards in Eqs.~\eqref{eq:m3-reward} and~\eqref{eq:m4-reward} with PPO~\cite{schulman2017proximal}. 
The actor observes the VLA latent feature $h_t$. 
Simulator state is used to compute the object-state rewards and to train the critic, but it is not provided to the deployed actor. 
The reward compiler uses object-state reward potentials from a small library $\mathcal{R}=\{\varphi_k\}_{k=1}^{K}$. 
In our experiments, the primitive components are distance progress, grasp state, placement progress, and articulation closing progress:
\begin{equation}
\begin{aligned}
\varphi_{\mathrm{dist}}(a,b)
&=
\max\left(-1,\;1-\frac{\|p(a)-p(b)\|_2}{d_{\mathrm{init}}(a,b)}\right),\\
\varphi_{\mathrm{grasp}}(\mathrm{gripper},\mathrm{obj})
&=
q_{\mathrm{grasp}}(\mathrm{gripper},\mathrm{obj})\in[0,1],\\
\varphi_{\mathrm{place}}(\mathrm{obj},\mathrm{target})
&=
\max\left(-1,\;1-
\frac{\|p(\mathrm{obj})-p(\mathrm{target})\|_2}
{d_{\mathrm{init}}(\mathrm{obj},\mathrm{target})}
\right),\\
\varphi_{\mathrm{close}}(\mathrm{obj})
&=
\max\left(-1,\;1-
\frac{|\theta(\mathrm{obj})-\theta_{\mathrm{closed}}|}
{|\theta_{\mathrm{init}}(\mathrm{obj})-\theta_{\mathrm{closed}}|}
\right).
\end{aligned}
\end{equation}
Where
$p(\cdot)$ denotes Cartesian position, $\theta(\cdot)$ denotes the relevant articulation state, $\theta_{\mathrm{closed}}$ is the closed target angle, and $q_{\mathrm{grasp}}$ is the simulator grasp-quality or grasp-indicator signal. 
The normalizers $d_{\mathrm{init}}(\cdot,\cdot)$ and $\theta_{\mathrm{init}}(\cdot)$ are measured when recovery is triggered. 

For a residual method with reward $R^{u}$, where $u$ denotes either the task instruction $\ell$ for M2 or the failure category $c$ for M3/M4/M6, the training objective is:
\begin{equation}
\theta_u^\star
=
\arg\max_{\theta_u}
\mathbb{E}
\left[
\sum_{t=0}^{T_u-1}
\gamma^t
R^{u}(s_t,a_t,s_{t+1})
\right],
\end{equation}
with actions composed as in Eq.~\eqref{eq:residual-action}. 
After training, the learned residual policies are frozen and stored in a recovery policy library. 
More details of the training can be found in Appendix~\ref {app:training-details}.

\section{Experimental Results}
\label{sec:result}

\subsection{Experiment setup}
We evaluate all six variants M1--M6 on a Fetch robot in both simulation and physical experiments.
The benchmark includes three tabletop tasks spanning distinct recovery regimes: contact-rich toolbox organization, long-horizon vegetable sorting, and short-horizon soda-can disposal.
Each task is specified by a language instruction and evaluated over 20 trials.
We report both binary success and a normalized goal-fulfillment quality score (Q-score), which assigns partial credit for intermediate recovery progress.
Task prompts, Q-score rubrics, and additional experimental details are provided in Appendix~\ref{app:setup-details}.
Figure~\ref{fig:experiment-examples} shows representative task setups, detected failure states, and recovered states.

Across simulation rollouts, we observe several recurring recoverable failures.
In the toolbox task, common failures include a cable or box left on the ground, an object grasped but not inserted into the toolbox, and a toolbox lid that remains open after object placement.
In the sorting task, failures include vegetables placed in the wrong receptacle, objects remaining on the table, and incomplete category-specific placement.
In the soda-can disposal task, failures include a can remaining on the table, a dropped can near the bin, and a can held by the gripper but not released into the trash can. A complete library of trained residual policies is provided in Appendix~\ref{app:setup-details}.

The mixture of residual policies library is deployed zero-shot on the physical robot. 
The robot executes the frozen base VLA by default. 
In parallel, the VLM maintains a sparse history of five RGB observations: $ H_t = \{I_{t-4\Delta},I_{t-3\Delta},I_{t-2\Delta},I_{t-\Delta},I_t\}, \Delta=15\mathrm{s}$. 
Given this history, the instruction, and the recovery prompt, the VLM works as failure state detector, and residual recovery is activated only when the VLM detects a failure category that
was observed and trained in simulation:
    \begin{equation}
    \beta_t
    =
    \mathbf{1}
    \left[
    \rho_t\ge \tau_{\mathrm{deploy}}
    \;\wedge\;
     c_t\in\mathcal{C}_{\mathrm{train}}
    \right].
    \end{equation}
If $\beta_t=1$, the dispatcher invokes the trained residual policy associated
with the detected category. 
If $c_t\notin\mathcal{C}_{\mathrm{train}}$, no
residual policy is called; the system continues with the base VLA, and the trial
is counted as unrecovered if the base policy cannot recover from that state.

\subsection{Simulation and physical results}
\begin{figure}[tbp]
    \centering
    \includegraphics[width=\linewidth]{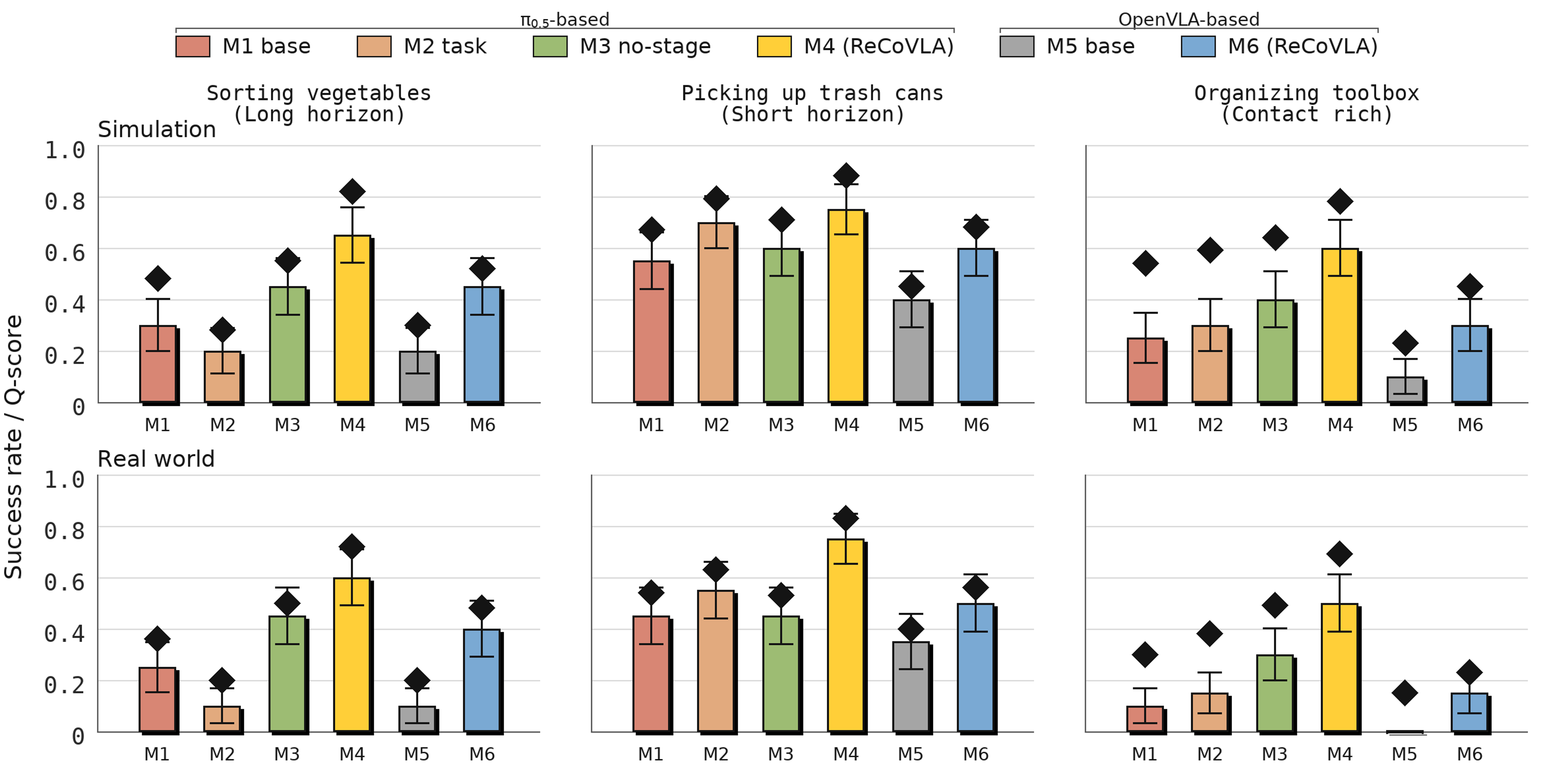}
    \vspace{-10pt}
    \caption{Simulation and physical experiments over 20 trials per method and task. The top and bottom rows report simulation and physical results, respectively. Bars show success rate, error bars show binomial standard error, and black diamonds show Q-score.}
    \label{fig:sim-physical-results}
    \vspace{-10pt}
\end{figure}

Figure~\ref{fig:sim-physical-results} summarizes the simulation and physical results. 
We observe that the full stage-gated reward compiler, M4, achieves the strongest overall performance. 
In simulation, M4 improves average success from \(36.7\%\) for the no-recovery $\pi_{0.5}$ baseline M1 to \(66.7\%\), with average Q-score increasing from 0.56 to 0.83. 
In physical experiments, M4 reaches \(61.7\%\) average success and 0.75 average Q-score, outperforming the baselines on each task by 18.3 points in success and 0.21 in Q-score.

The ablations indicate that recovery depends strongly on reward structure. 
M2 uses a task-level residual reward, while M3 activates VLM-selected failure-relevant reward components without stage-aware gates. 
Both improve some tasks, but neither matches M4. 
In simulation, M3 reaches \(48.3\%\) average success, 18.4 percentage points below M4, and has lower Q-scores on every task. 
Thus, selecting relevant reward terms is useful but insufficient: the main gain comes from stage-aware gates that suppress reward components before their recovery preconditions are satisfied.

Task-level results support the same conclusion. 
In simulation, M4 achieves the best result on all three tasks, improving toolbox organization from \(25\%\) to \(60\%\) success and vegetable sorting from \(30\%\) to \(65\%\). 
These gains are largest in tasks requiring staged recovery, such as re-grasping, object placement, and lid closing. 
The same design also benefits another VLA backbone: M6 improves OpenVLA average success from \(23.3\%\) for M5 to \(45.0\%\). 
In physical experiments, M4 again performs best on all tasks, reaching \(50\%\), \(60\%\), and \(75\%\) success on toolbox organization, vegetable sorting, and soda-can disposal, respectively. 
The soda-can task is the most favorable to generic recovery, with M2 reaching \(55\%\) success, but M4 still achieves the highest success and Q-score. Overall, M1 and M5 show that base VLAs remain brittle after failures, while the M3--M4 gap isolates the value of stage-gated reward compilation over simply activating failure-relevant rewards. 

\subsection{Out of distribution stress test}

\begin{wrapfigure}[10]{r}{0.55\linewidth}
    \centering
    \setlength{\tabcolsep}{1pt}
    \begin{tabular}{@{}ccc@{}}
        \includegraphics[width=0.30\linewidth]{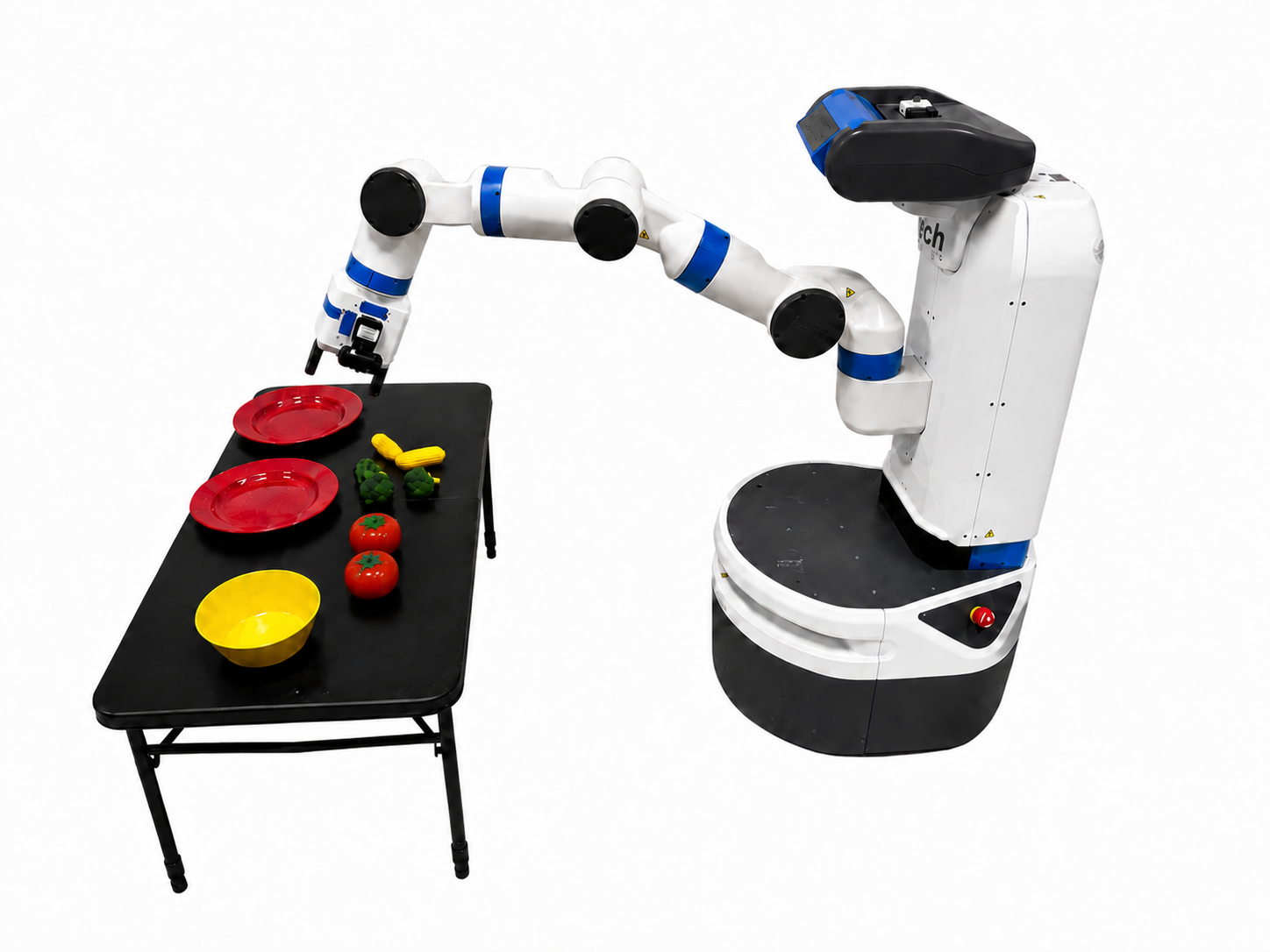} &
        \includegraphics[width=0.30\linewidth]{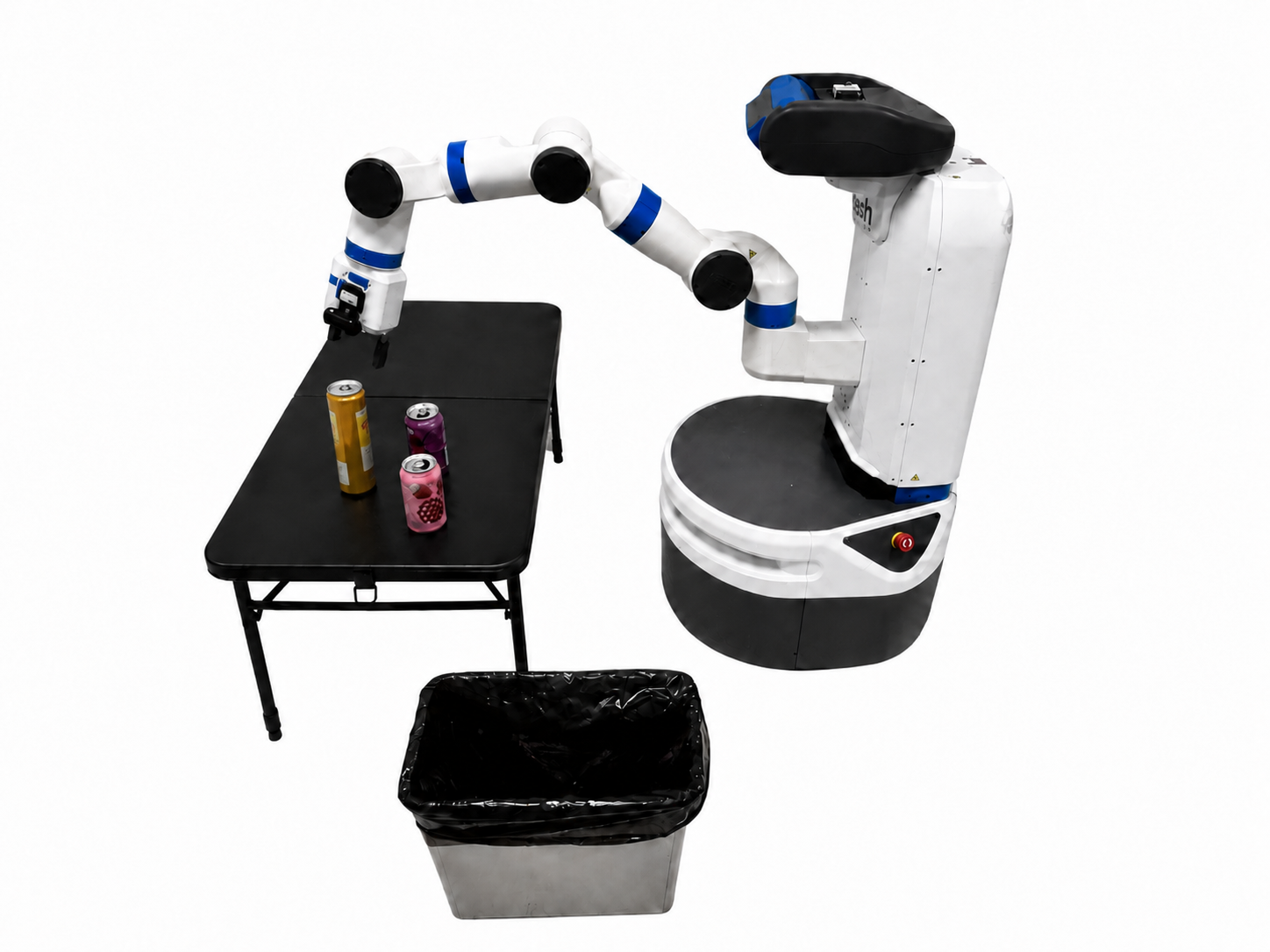} &
        \includegraphics[width=0.30\linewidth]{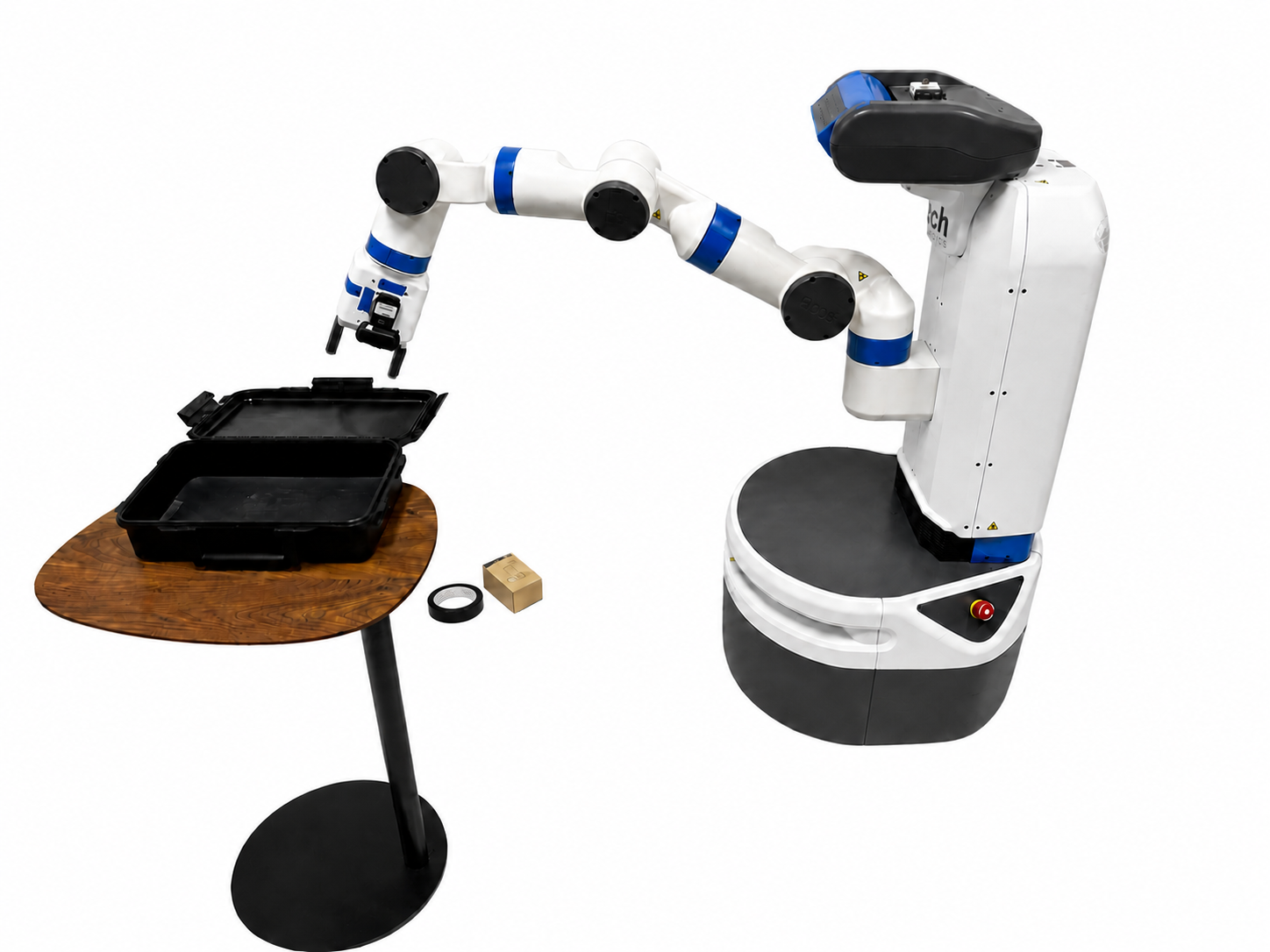} \\
        \small Sort & \small Dispose & \small Organize \\
    \end{tabular}

    \small
    \setlength{\tabcolsep}{2.0pt}
    \renewcommand{\arraystretch}{0.88}
    \begin{tabular}{lccc}
        \toprule
        Task & OOD & M1 & M4 \\
        \midrule
        Sort & pumpkin \(\rightarrow\) tomato & \(0\%\)/0.10 & \(50\%\)/0.65 \\
        Dispose & tall can & \(30\%\)/0.50 & \(70\%\)/0.76 \\
        Organize & cable \(\rightarrow\) tape & \(0\%\)/0.05 & \(40\%\)/0.55 \\
        \addlinespace
        Avg. & --- & \(10.0\%\)/0.22 & \(53.3\%\)/0.65 \\
        \bottomrule
    \end{tabular}
    \caption{OOD setups and results in success/Q-score.}
    \label{fig:ood-results}
\end{wrapfigure}

Figure~\ref{fig:ood-results} reports an out-of-distribution (OOD) evaluation under object substitutions that preserve task structure but change visual appearance or geometry. 
The no-recovery base policy M1 drops to \(10.0\%\) average success and 0.22 Q-score, including zero successful trials for two tasks. 
Under the same OOD changes, M4 achieves \(53.3\%\) average success and 0.65 Q-score, with a 43.3 percentage-point success gain and a 0.43 Q-score gain over M1. 
The soda-can variant remains the most robust at \(70\%\) success and 0.76 Q-score for M4, while the contact-rich toolbox variant is harder. 

To validate the advantage of M4 over recent recovery and VLA+RL methods, we additionally evaluate against RLinf~\cite{yu2025rlinf} and RACER~\cite{dai2024racer}, and test M4 on Behavior-1K~\cite{li2024behavior1k} using a different mobile humanoid manipulator, the Galaxea R1 Pro. 
More details are provided in Appendix~\ref{app:additional-results}.

\vspace{6pt}
\section{Conclusion}
\label{sec:conclusion}
\vspace{6pt}
We presented a failure-conditioned reward compilation framework for residual VLA recovery. 
Instead of fine-tuning the full base policy or training a residual policy with a fixed reward, \method keeps the pretrained VLA frozen, uses an external VLM to infer the current failure mode and recovery stage, and compiles a reward from selected task-grounded components. 
This design lets the residual policy receive feedback matched to the current recovery subproblem while preserving the nominal competence of the base policy. 
Across three manipulation tasks, the experiments show that reward selection is central to reliable recovery. 
The proposed \method is the strongest method overall, increasing average simulation success from \(36.7\%\) for the fine-tuned $\pi_{0.5}$ baseline to \(66.7\%\). 
The same stage-gated recovery mechanism also improves OpenVLA in simulation and physical experiments. 
In physical experiments, \method achieves the best success and Q-score on all three tasks and outperforms the other tested recovery methods on average. 
These results suggest that semantic failure understanding should shape not only the choice to recover, but also the reward structure used to learn the recovery behavior.

\section{Limitations}
\label{sec:limitation}
\method requires each recoverable failure category to be reproducible in simulation before deployment. 
When a new failure mode is observed on the real robot, we can add it to the recovery catalog by reproducing the corresponding state distribution in simulation, compiling a stage-gated reward, and training a new residual policy. 
However, the current system does not synthesize or train a new recovery policy online during a physical trial. 
The empirical evaluation is limited in scale. 
We evaluate three tabletop Fetch manipulation tasks with 20 trials per task and method, covering long-horizon sorting, short-horizon disposal, and contact-rich toolbox organization, but not the full diversity of manipulation failures. 
Residual policies are trained in simulation and deployed zero-shot, so performance can still be affected by sim-to-real mismatch, perception errors, contact variation, and VLM failure-classification mistakes. 
Future work should automate failure-catalog expansion, learn or generate reward gates, and evaluate broader physical baselines.

\clearpage

\bibliography{main}  %

\clearpage
\appendix
\setcounter{topnumber}{5}
\setcounter{bottomnumber}{5}
\setcounter{totalnumber}{8}
\renewcommand{\topfraction}{0.95}
\renewcommand{\bottomfraction}{0.85}
\renewcommand{\textfraction}{0.07}
\renewcommand{\floatpagefraction}{0.80}

\section{Appendix: Additional Experimental Details}
\label{sec:appendix}

\subsection{Additional results}
\label{app:additional-results}

\begin{figure}[H]
    \centering
    \includegraphics[width=0.96\linewidth]{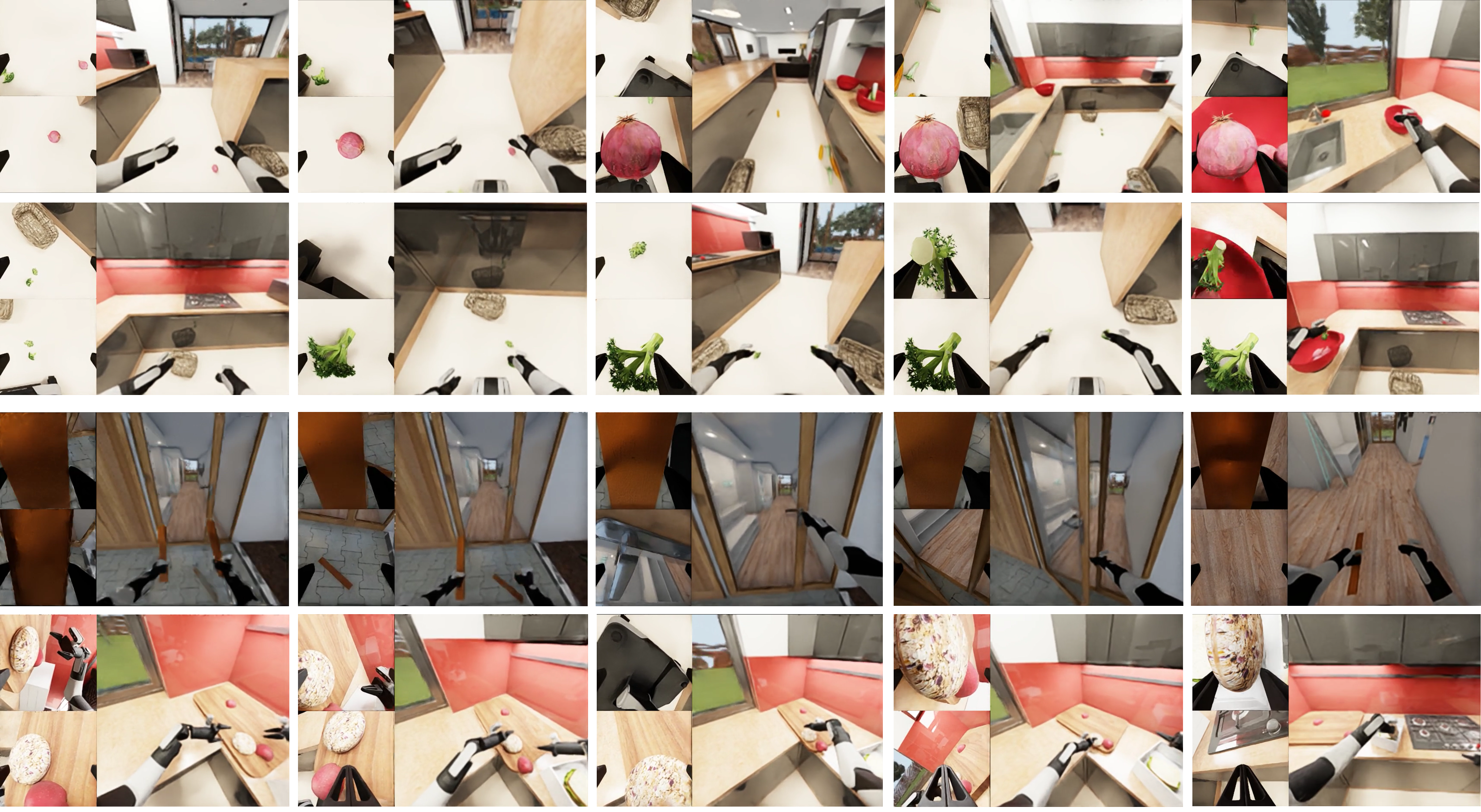}
    \caption{Failure recovery examples on Behavior-1K Challenge tasks. The top two rows show recovery from dropped onion and broccoli failures: the robot re-grasps the object from the ground and places it into the red bowl. The third row shows a door-opening failure in which both hands are occupied by wood sticks; the learned recovery drops one stick, opens the door, and continues the task. The final row shows recovery for picking up a bagel and placing it into the lunch box.}
    \label{fig:b1k}
\end{figure}

\paragraph{Test on Behavior-1k challenge.}
To evaluate whether the recovery design can be transferred beyond the Fetch tabletop setup, we additionally implemented \method on three tasks in the Behavior-1K Challenge\footnote{B1K Challenge: \url{https://behavior.stanford.edu/index.html}}: sorting vegetables, bringing in wood, and preparing a lunch box. 
These tasks use a different simulated robot platform (R1 Pro) and require recovery from object drops, constrained bimanual manipulation, and placement errors. Figure~\ref{fig:b1k} shows representative recovery trajectories. Across these examples, the VLM descriptor identifies the failed object or manipulation stage, while the compiled residual reward encourages the local correction needed to return the rollout to a solvable state. Table~\ref{tab:b1k_test} shows the performance of \method and Openpi Comet\footnote{A top-ranking VLA model in the B1K Challenge: \url{https://github.com/mli0603/openpi-comet}}, demonstrating the advantage of our method.

\begin{table}[H]
\centering
\caption{Behavior-1K challenge test results. Each method is evaluated over 20 trials per task. Success rate and Q-score are reported in $[0,1]$, where higher is better.}
\label{tab:b1k_test}
\small
\setlength{\tabcolsep}{5pt}
\renewcommand{\arraystretch}{1.10}
\begin{tabular}{llccc}
\toprule
Task & Method & Trials & Success rate & Q-score \\
\midrule
Sorting vegetables on the table & ReCoVLA & 20 & \textbf{0.20} & \textbf{0.51} \\
Sorting vegetables on the table & OpenPI Comet & 20 & 0.05 & 0.26 \\
\midrule
Bringing in wood & ReCoVLA & 20 & \textbf{0.65} & \textbf{0.79} \\
Bringing in wood & OpenPI Comet & 20 & 0.50 & 0.62 \\
\midrule
Preparing a lunch box & ReCoVLA & 20 & \textbf{0.25} & \textbf{0.49} \\
Preparing a lunch box & OpenPI Comet & 20 & 0.10 & 0.33 \\
\midrule
Average over three tasks & ReCoVLA & -- & \textbf{0.37} & \textbf{0.60} \\
Average over three tasks & OpenPI Comet & -- & 0.22 & 0.40 \\
\bottomrule
\end{tabular}
\end{table}

\paragraph{Comparison with additional baselines.} 
We also compare \method with two recent recovery or VLA+RL baselines, RLinf~\cite{yu2025rlinf} and RACER~\cite{dai2024racer}, on the same three simulation tasks. Table~\ref{tab:additional_baselines} reports 20 trials per method and task. Averaged over the three tasks, \method obtains 0.67 success and 0.83 Q-score, compared with 0.12/0.19 for RLinf and 0.43/0.60 for RACER. RACER performs reasonably on the short-horizon soda-can task, but remains lower than \method on all three tasks. RLinf is more sensitive to the sparse and staged nature of the recovery objectives, particularly in vegetable sorting and toolbox organization. These comparisons support the main result that failure-conditioned, stage-gated reward compilation is important for robust recovery rather than merely adding an RL correction to the base VLA.

\begin{table}[!tbp]
\centering
\caption{Additional comparison with existing baselines. Each method is evaluated over 20 trials per task. Success rate and Q-score are reported in $[0,1]$, where higher is better.}
\label{tab:additional_baselines}
\small
\setlength{\tabcolsep}{5pt}
\renewcommand{\arraystretch}{1.10}
\begin{tabular}{llccc}
\toprule
Task & Method & Trials & Success rate & Q-score \\
\midrule
Sorting vegetables & ReCoVLA & 20 & \textbf{0.65} & \textbf{0.82} \\
Sorting vegetables & RLinf & 20 & 0.00 & 0.12 \\
Sorting vegetables & RACER & 20 & 0.35 & 0.49 \\
\midrule
Soda-can disposal & ReCoVLA & 20 & \textbf{0.75} & \textbf{0.88} \\
Soda-can disposal & RLinf & 20 & 0.35 & 0.40 \\
Soda-can disposal & RACER & 20 & 0.65 & 0.74 \\
\midrule
Organizing toolbox & ReCoVLA & 20 & \textbf{0.60} & \textbf{0.78} \\
Organizing toolbox & RLinf & 20 & 0.00 & 0.05 \\
Organizing toolbox & RACER & 20 & 0.30 & 0.56 \\
\midrule
Average over three tasks & ReCoVLA & -- & \textbf{0.67} & \textbf{0.83} \\
Average over three tasks & RLinf & -- & 0.12 & 0.19 \\
Average over three tasks & RACER & -- & 0.43 & 0.60 \\
\bottomrule
\end{tabular}
\end{table}

\subsection{Additional experiment setup details}
\label{app:setup-details}
\paragraph{Additional physical experiment details.}
All physical experiments use a Fetch mobile manipulator with the task prompts listed in Table~\ref{tab:tasks}. In the vegetable-sorting task, we randomize vegetable positions on the table while keeping the yellow bowl and red dishes fixed. In the soda-can disposal task, we randomize soda-can positions and appearance while keeping the can size fixed. In the toolbox organization task, we randomize the cable and camera-box positions on the ground. These randomizations preserve the task semantics while changing the initial recovery geometry across trials.

During physical experiments, we terminate a trial if the robot makes no measurable progress for a 3-minute window, defined as no increase in Q-score over that interval. We also terminate a trial if the robot collides with the environment or an undesired object. The maximum execution time for each task is set to twice the average duration of the corresponding human expert demonstrations.

\paragraph{Q-score criteria.}
For vegetable sorting, each pumpkin placed in the yellow bowl receives 0.15, each broccoli piece placed in the middle red dish receives 0.10, and each corn placed in the right red dish receives 0.15; full task completion receives a Q-score of 1.0. For soda-can disposal, each can placed in the trash bin receives 0.30, with full completion assigned 1.0. For toolbox organization, placing the cable or box in the toolbox receives 0.30 each, and closing the lid receives 0.40; full completion receives 1.0.

\begin{table}[!tbp]
    \centering
    \small
    \caption{Evaluation tasks and language prompts.}
    \label{tab:tasks}
    \begin{tabular}{p{0.20\linewidth}p{0.18\linewidth}p{0.52\linewidth}}
        \toprule
        Task & Purpose & Language prompt \\
        \midrule
        Organizing toolbox &
        Contact rich &
        Pick up the cable and box on the ground and place them in the toolbox
        on the table, and close the lid of the toolbox. \\
        Sorting vegetables &
        Long horizon &
        Sort the vegetables into the red dishes and yellow bowl on the table:
        put all three broccoli pieces into one red dish; put two corns into the
        other red dish; and put both pumpkin pieces into the yellow bowl. \\
        Picking up trash cans &
        Short horizon &
        Put the three cans of soda from the table inside the trash can on the
        left of the table. \\
        \bottomrule
    \end{tabular}
\end{table}

\paragraph{Simulation environment.} The simulation environments mirror the three layouts used in the physical Fetch experiments. The toolbox environment contains a table-mounted toolbox and ground objects that must be picked up, inserted into the toolbox, and followed by lid closure. The sorting environment contains red dishes, a yellow bowl, and multiple vegetable categories that must be placed into their specified containers. The trash-can environment contains soda cans on the table and a trash can placed beside the table. These layouts instantiate the same recovery regimes shown in Figure~\ref{fig:experiment-examples}: contact-rich recovery, long-horizon object-category sorting, and short-horizon disposal.

During simulation rollouts, the external VLM failure analyzer is queried on the current RGB observation and recovery prompt. The resulting descriptor records the failure type, recovery stage, active entities, confidence, and reward mask. In the toolbox task, typical detected failure states include a cable or box left on the ground, an object grasped but not inserted into the toolbox, and a toolbox lid that remains open after object placement. In the sorting task, typical failures include vegetables placed in the wrong receptacle, remaining objects on the table, and incomplete category-specific placement. In the trash-can task, typical failures include a can remaining on the table, a dropped can near the bin, or a can held by the gripper but not yet released into the trash can. These detected states determine which entities and reward-library components are activated by the reward compiler during residual policy optimization.

\subsection{Reward compiler implementation and VLM failure detector confusion matrix}
\label{app:reward-details}

\paragraph{Implementation details of reward compiler.} 
Algorithm~\ref{alg:reward-compiler} gives the implementation-level reward compilation procedure used in our experiments. Beyond the high-level description in the main paper, the implementation performs conservative validation before any residual reward is produced. It normalizes the VLM category and stage labels to the task vocabulary, checks that the reward mask has length \(K\), rejects descriptors with unresolved entities, and rejects descriptors whose selected components cannot be bound to the required semantic roles. This rejection behavior is intentional: if the descriptor is incomplete or inconsistent with the simulator object map, the system does not synthesize a fallback reward. Instead, the dispatcher leaves the nominal VLA active or invokes no recovery policy for that descriptor. The algorithm also makes the component-level binding explicit. The active entity set \(E_c\) may contain extra objects observed by the VLM, but each reward potential \(\varphi_k\) receives only the entities required by its signature \(\sigma_k\). This avoids, for example, passing a receptacle entity to a grasp term or an end-effector entity to an articulation term. Stage gates are retrieved only after role binding, so each gate is parameterized by the same grounded entities used by the corresponding reward term.

\begin{algorithm}[!tbp]
\caption{Deterministic reward compilation}
\label{alg:reward-compiler}
\begin{algorithmic}[1]
\Require VLM descriptor $\xi_c=(c,z,E_c,\rho,m_c)$; task specification 
$\mathcal{T}$; simulator object map $\mathcal{O}$; reward library 
$\mathcal{R}=\{\varphi_k\}_{k=1}^K$; component signatures $\{\sigma_k\}_{k=1}^K$; 
stage-gate templates $\mathcal{G}$; recoverable catalog $\mathcal{C}_{\mathrm{train}}$; confidence threshold $\tau$
\Ensure Executable reward $R_{\mathrm{M4}}^{c}(s_t,a_t,s_{t+1})$, or rejection

\State Normalize category and stage labels:
\Statex \hspace{\algorithmicindent}
$(\hat{c},\hat{z}) \gets \textsc{NormalizeLabels}(c,z,\mathcal{T})$

\If{$\hat{c}\notin \mathcal{C}_{\mathrm{train}}$ \textbf{or} $\rho < \tau$ \textbf{or} $m_c\notin\{0,1\}^K$}
    \State Reject descriptor and return no recovery reward.
\EndIf

\State Canonicalize VLM entities:
\Statex \hspace{\algorithmicindent}
$\hat{E}_c \gets \textsc{ResolveEntities}(E_c,\mathcal{T},\mathcal{O})$

\If{$\hat{E}_c$ contains unresolved entities}
    \State Reject descriptor and return no recovery reward.
\EndIf

\State Assign semantic roles:
\Statex \hspace{\algorithmicindent}
$\mathcal{R}_c \gets \textsc{AssignRoles}(\hat{E}_c,\hat{c},\hat{z},\mathcal{T})$
\Statex \hspace{\algorithmicindent}
(e.g., end-effector, object, target, source, articulated part)

\State Initialize selected term set $\mathcal{S}_c \gets \emptyset$.

\For{$k=1,\ldots,K$}
    \If{$m_c^{(k)}=1$}
        \State Read the component signature $\sigma_k$ for $\varphi_k$.
        \State Bind component-specific entities:
        \Statex \hspace{\algorithmicindent}
        $\bar{E}_{c,k} \gets \textsc{BindRoles}(\mathcal{R}_c,\sigma_k)$

        \If{$\bar{E}_{c,k}$ is missing any required role}
            \State Reject descriptor and return no recovery reward.
        \EndIf

        \State Retrieve the fixed stage gate:
        \Statex \hspace{\algorithmicindent}
        $g_{\hat{c},k}(\cdot) \gets \mathcal{G}(\hat{c},\hat{z},k,\bar{E}_{c,k})$

        \State Add the gated potential-difference term:
        \Statex \hspace{\algorithmicindent}
        $\mathcal{S}_c \gets \mathcal{S}_c \cup \left\{ g_{\hat{c},k}(s_t;\bar{E}_{c,k}) \Delta\varphi_k(s_t,s_{t+1};\bar{E}_{c,k}) \right\}$
    \EndIf
\EndFor

\If{$\mathcal{S}_c=\emptyset$}
    \State Reject descriptor and return no recovery reward.
\EndIf

\State Return the compiled reward:
\Statex \hspace{\algorithmicindent}
$R_{\mathrm{M4}}^{c}(s_t,a_t,s_{t+1}) = \sum_{\psi \in \mathcal{S}_c} \psi - \Omega_t(a_t^r)$
\end{algorithmic}
\end{algorithm}

\paragraph{Quantitative results of experiments.} 
Table~\ref{tab:failure_recovery_all_results} provides the full numerical results corresponding to the main-paper plots, including simulation, physical robot, and OOD settings. The averages show that M4, the full \method instantiation on $\pi_{0.5}$, is the strongest variant in both simulation and physical deployment. In simulation, M4 improves the average result from 0.37/0.56 for the no-recovery $\pi_{0.5}$ baseline M1 to 0.67/0.83. On the physical robot, M4 improves the average from 0.27/0.40 to 0.62/0.75. The OOD rows show the same trend under object substitutions: M4 achieves 0.53/0.65 on average, while M1 reaches only 0.10/0.22. The OpenVLA comparison further indicates that the recovery design is not tied to a single base VLA, with M6 improving over M5 in both simulation and physical experiments.

\begin{table}[!tbp]
\centering
\caption{Failure-recovery experiment results in simulation, physical robot, and out-of-distribution (OOD) settings. Each entry reports success rate / Q-score. Higher is better.}
\label{tab:failure_recovery_all_results}
\begin{threeparttable}
\scriptsize
\setlength{\tabcolsep}{3.5pt}
\renewcommand{\arraystretch}{1.15}
\resizebox{\linewidth}{!}{%
\begin{tabular}{llcccccc}
\toprule
Setting & Task / split & M1 & M2 & M3 & M4 (Ours) & M5 & M6 \\
\midrule
\multicolumn{8}{l}{\textit{Average over three in-distribution tasks}} \\
Simulation & Average & 0.37 / 0.56 & 0.40 / 0.55 & 0.48 / 0.63 & \textbf{0.67 / 0.83} & 0.23 / 0.33 & 0.45 / 0.55 \\
Physical robot & Average & 0.27 / 0.40 & 0.27 / 0.40 & 0.40 / 0.51 & \textbf{0.62 / 0.75} & 0.15 / 0.25 & 0.35 / 0.42 \\
\midrule
\multicolumn{8}{l}{\textit{In-distribution task results}} \\
Simulation & Sorting vegetables & 0.30 / 0.48 & 0.20 / 0.28 & 0.45 / 0.55 & \textbf{0.65 / 0.82} & 0.20 / 0.30 & 0.45 / 0.52 \\
Simulation & Picking up trash cans & 0.55 / 0.67 & 0.70 / 0.79 & 0.60 / 0.71 & \textbf{0.75 / 0.88} & 0.40 / 0.45 & 0.60 / 0.68 \\
Simulation & Organizing toolbox & 0.25 / 0.54 & 0.30 / 0.59 & 0.40 / 0.64 & \textbf{0.60 / 0.78} & 0.10 / 0.23 & 0.30 / 0.45 \\
\addlinespace[2pt]
Physical robot & Sorting vegetables & 0.25 / 0.36 & 0.10 / 0.20 & 0.45 / 0.50 & \textbf{0.60 / 0.72} & 0.10 / 0.20 & 0.40 / 0.48 \\
Physical robot & Picking up trash cans & 0.45 / 0.54 & 0.55 / 0.63 & 0.45 / 0.53 & \textbf{0.75 / 0.83} & 0.35 / 0.40 & 0.50 / 0.56 \\
Physical robot & Organizing toolbox & 0.10 / 0.30 & 0.15 / 0.38 & 0.30 / 0.49 & \textbf{0.50 / 0.69} & 0.00 / 0.15 & 0.15 / 0.23 \\
\midrule
\multicolumn{8}{l}{\textit{OOD stress tests}} \\
OOD (physical robot) & \makecell[l]{Sorting vegetables\\pumpkin $\rightarrow$ tomato} & 0.00 / 0.10 & -- & -- & \textbf{0.50 / 0.65} & -- & -- \\
OOD (physical robot) & \makecell[l]{Picking up trash cans\\soda can $\rightarrow$ tall soda can} & 0.30 / 0.50 & -- & -- & \textbf{0.70 / 0.76} & -- & -- \\
OOD (physical robot) & \makecell[l]{Organizing toolbox\\cable $\rightarrow$ tape} & 0.00 / 0.05 & -- & -- & \textbf{0.40 / 0.55} & -- & -- \\
\bottomrule
\end{tabular}%
}
\begin{tablenotes}[flushleft]
\footnotesize
\item M1: fine-tuned $\pi_{0.5}$ without recovery. M2: task-level residual reward. M3: equal-weight failure rewards. M4: proposed $\pi_{0.5}$ VLM reward compiler. M5: fine-tuned OpenVLA without recovery. M6: OpenVLA with the proposed recovery method.
\item Bold entries denote the best result in each row. OOD tests compare M4 against the corresponding non-recovery baseline M1.
\end{tablenotes}
\end{threeparttable}
\end{table}

\begin{figure}[tbp]
    \centering
    \includegraphics[width=0.85\linewidth]{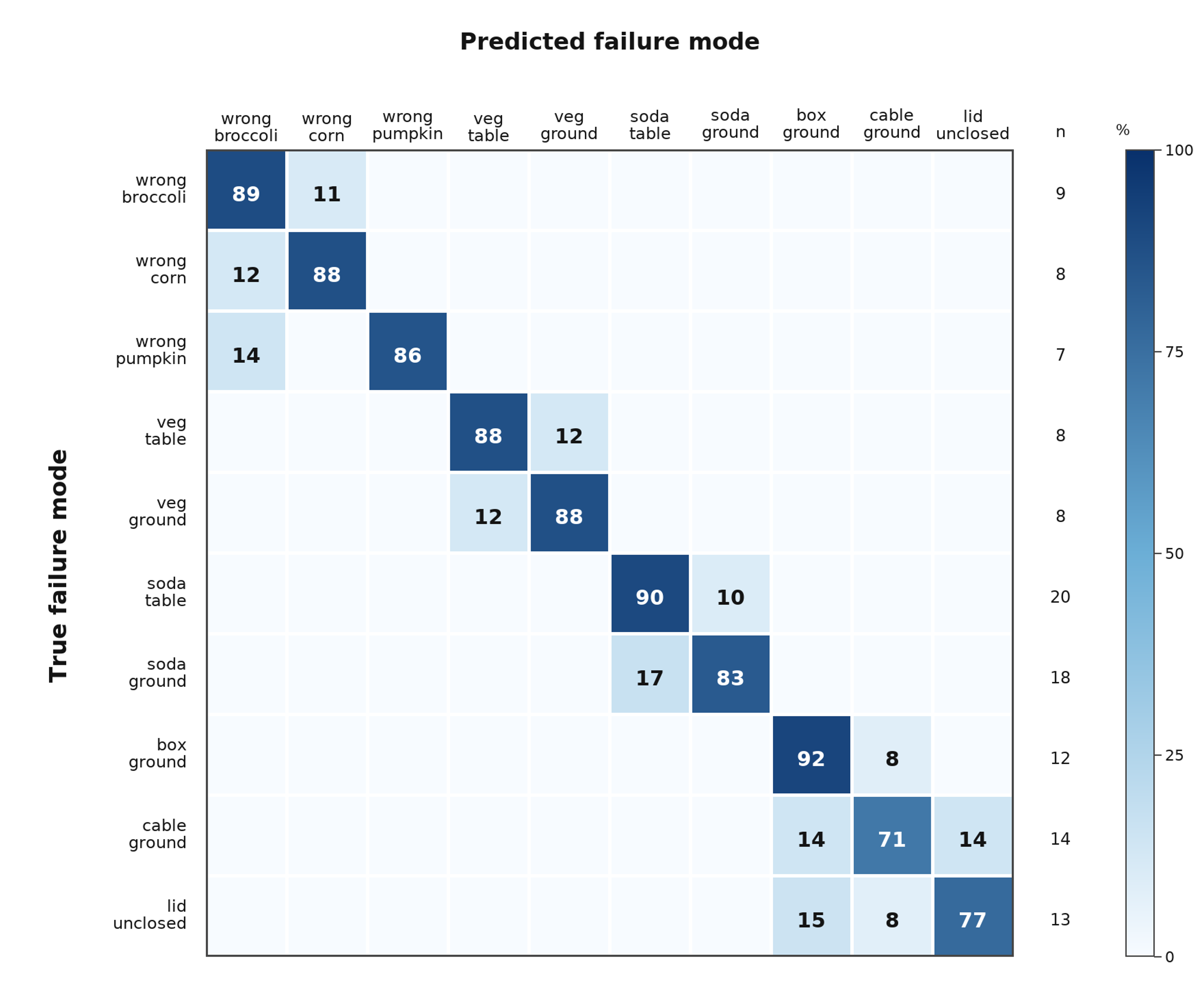}
    \caption{VLM failure detector confusion matrix. Rows are normalized by the true failure mode, and the support column reports the number of physical-rollout occurrences used for each row.}
    \label{fig:confusion-matrix}
\end{figure}

\paragraph{VLM failure-analysis accuracy.}
Figure~\ref{fig:confusion-matrix} reports a row-normalized failure-mode confusion matrix for the external VLM detector used during physical rollouts. The matrix covers ten recoverable failure modes: five from vegetable sorting, two from soda-can disposal, and three from toolbox organization. Because the active task is known from the language instruction, the evaluation constrains predictions to the corresponding task group; errors therefore measure confusion between semantically related failures within the same task, rather than impossible cross-task predictions. For vegetable sorting, the observed failure occurrences are 9 wrong-broccoli-receptacle, 8 wrong-corn-receptacle, 7 wrong-pumpkin-receptacle, 8 dropped-vegetable-on-table, and 8 dropped-vegetable-on-ground cases, totaling 40 detections with 87.5\% aggregate accuracy, 87.4\% macro accuracy, and 1.0 percentage-point class-wise standard deviation. For soda-can disposal, the detector observes 20 fallen-can-on-table and 18 dropped-can-on-ground cases, totaling 38 detections with 86.8\% aggregate accuracy, 86.7\% macro accuracy, and 3.3 percentage-point standard deviation. For toolbox organization, the detector observes 12 dropped-box, 14 dropped-cable, and 13 unclosed-lid cases, totaling 39 detections with 79.5\% aggregate accuracy, 80.0\% macro accuracy, and 8.5 percentage-point standard deviation. Across all tasks, the detector reaches 85.0\% macro accuracy, 84.6\% aggregate accuracy, and 6.0 percentage-point class-wise standard deviation. Most errors are local, such as confusion between wrong vegetable receptacles, table-versus-ground soda-can failures, or cable/box/lid failures in the toolbox task. This supports the use of the VLM as a semantic failure dispatcher, while also highlighting that detector mistakes can still select a suboptimal residual policy in visually similar recovery states.

\subsection{Training details}
\label{app:training-details}
\paragraph{Base VLA fine-tuning.}
For each task, we collected 40 expert demonstrations: 20 in simulation and 20 on the physical robot. We use these demonstrations to fine-tune the $\pi_{0.5}$ base model used by M1--M4. The fine-tuned policy is then used as the nominal controller for all variants built on the same backbone. During residual recovery training, the base VLA is frozen, so no gradients are propagated into the VLA encoder, language model, or action expert. The same process is applied to fine-tune the OpenVLA base. The main fine-tuning parameters are listed in Table~\ref{tab:pi05-finetune-params}.

\begin{table}[!tbp]
\centering
\caption{Training parameters for base-VLA fine-tuning and residual PPO.}
\label{tab:pi05-finetune-params}
\label{tab:residual-ppo-params}
\scriptsize
\setlength{\tabcolsep}{4pt}
\renewcommand{\arraystretch}{1.05}
\begin{tabular}{lll}
\toprule
Stage & Parameter & Value \\
\midrule
\multicolumn{3}{l}{\textit{$\pi_{0.5}$ fine-tuning}} \\
\midrule
$\pi_{0.5}$ & Expert demonstrations per task & 40 total: 20 simulation, 20 physical robot \\
$\pi_{0.5}$ & Base checkpoint & \texttt{gs://openpi-assets/checkpoints/pi05\_base/params} \\
$\pi_{0.5}$ & Action horizon & 32 \\
$\pi_{0.5}$ & Training steps & 20000 gradient steps \\
$\pi_{0.5}$ & Batch size & 4 \\
$\pi_{0.5}$ & Learning-rate schedule & Cosine decay \\
$\pi_{0.5}$ & Peak learning rate & \(2.5\times 10^{-5}\) \\
$\pi_{0.5}$ & Warmup steps & \(0.1\times\) training steps, capped at 1000 \\
$\pi_{0.5}$ & Seed & 42 \\
$\pi_{0.5}$ & Freeze rule & OpenPI $\pi_{0.5}$ model freeze filter \\
\midrule
\multicolumn{3}{l}{\textit{Residual PPO}} \\
\midrule
PPO & Policy input & Frozen $\pi_{0.5}$ VLA latent feature \(h_t\) \\
PPO & Actor/critic architecture & MLP, hidden sizes \((4096,1024,128)\) \\
PPO & Activation & Tanh \\
PPO & Total timesteps per residual policy & 300,000 \\
PPO & Vectorized environments & 1 \\
PPO & Episode horizon & 1000 steps \\
PPO & Simulator steps per residual action & 1 \\
PPO & PPO rollout length \(n_{\mathrm{steps}}\) & 100 \\
PPO & Discount factor \(\gamma\) & 0.99 \\
PPO & GAE parameter & 0.95 \\
PPO & PPO clip range & 0.2 \\
PPO & PPO learning rate & \(3\times 10^{-4}\) \\
PPO & Residual action clip & \([-1,1]\) before scaling \\
PPO & Default residual scale & 0.25 \\
PPO & Mobile-base residual scale & 0.03 \\
PPO & Torso residual scale & 0.01 \\
PPO & Startup gripper-open duration & 8 environment steps \\
PPO & Action-delta penalty coefficient & 0.005 \\
PPO & Action sign-flip penalty coefficient & 0.01 \\
\bottomrule
\end{tabular}
\end{table}

\paragraph{Residual policy training.}
Residual policies are trained entirely in simulation with PPO, using the same simulator task layouts and recoverable failure categories described in Section~\ref{app:setup-details}. For each recoverable failure category, rollouts are initialized from a restored OmniGibson scene state that reproduces the corresponding failure, such as an object in the wrong receptacle, a dropped soda can, a ground object near the toolbox, or an unclosed toolbox lid. The actor receives the frozen VLA latent feature \(h_t\) and outputs an additive residual action \(a_t^r\). The executed action is the clipped sum of the base action and residual action, as defined in Eq.~\ref{eq:residual-action}. After training, the residual actor is frozen and stored in the recovery policy library indexed by the VLM-detected failure category. PPO and residual-action parameters are listed in Table~\ref{tab:residual-ppo-params}.

\paragraph{Reward and regularization.}
The reward used for each residual rollout is generated by the deterministic compiler in Algorithm~\ref{alg:reward-compiler}. M2 uses a task-level reward mask, M3 uses the VLM-selected failure mask without stage gates, and M4/M6 use the same failure mask with stage-aware gates. The implementation also enforces per-joint action bounds from the simulator action space. In addition to the residual penalty \(\Omega_t(a_t^r)\), the implementation applies smoothness penalties on action deltas and sign-flip reversals, with coefficients reported in Table~\ref{tab:residual-ppo-params}. This regularization is important during sim-to-real transfer because it biases the residual policy toward small corrective motions around the nominal VLA behavior rather than replacing the base controller.

\paragraph{Deployment.}
The trained residual policies are deployed zero-shot on the physical Fetch robot. The robot executes the frozen base VLA by default. At a fixed monitoring interval, the VLM failure detector receives the recent RGB observation history and task prompt. If the detector predicts a known failure category with confidence above the deployment threshold, the corresponding residual policy is activated; otherwise the robot continues with the base policy. The residual policy is therefore used only as a targeted recovery controller for known failure states, not as a replacement for nominal task execution.

\end{document}